\documentclass[conference]{IEEEtran}
\IEEEoverridecommandlockouts
\usepackage{cite}
\usepackage{amsmath,amssymb,amsfonts}
\usepackage{algorithmic}
\usepackage{textcomp}
\usepackage{xcolor}
\usepackage{multirow}
\usepackage{adjustbox}

\usepackage{times}
\usepackage{epsfig}
\usepackage{gensymb}
\usepackage{float}
\usepackage{color}
\usepackage{booktabs}
\usepackage{lipsum} % for nonsense text
\usepackage{subfigure}

\def\BibTeX{{\rm B\kern-.05em{\sc i\kern-.025em b}\kern-.08em
    T\kern-.1667em\lower.7ex\hbox{E}\kern-.125emX}}

\begin{document}
\IEEEoverridecommandlockouts
\IEEEpubid{\makebox[\columnwidth]{978-1-6654-3288-7/21/\$31.00 \copyright 2021 European Union \hfill} 
%978-1-6654-3288-7/21/$31.00©2021 European Union
\hspace{\columnsep}\makebox[\columnwidth]{ }}

\title{Supervised Fine-tuning Evaluation for Long-term Visual Place Recognition}

\author{\IEEEauthorblockN{Farid Alijani}
\IEEEauthorblockA{\textit{Tampere University}\\
Tampere, Finland \\
farid.alijani@tuni.fi}

\and
\IEEEauthorblockN{Esa Rahtu}
\IEEEauthorblockA{\textit{Tampere University}\\
Tampere, Finland \\
esa.rahtu@tuni.fi}
}
\maketitle

\begin{abstract}
In this paper, we present a comprehensive study on the utility of 
deep convolutional neural networks with two state-of-the-art pooling
layers which are placed after convolutional layers and fine-tuned in 
an end-to-end manner for visual place recognition task in challenging
conditions, including seasonal and illumination variations. 
We compared extensively the performance of deep learned global features
with three different loss functions, e.g. triplet, contrastive and ArcFace,
for learning the parameters of the architectures in terms of fraction of the
correct matches during deployment. To verify effectiveness of our results,
we utilized two real world datasets in place recognition, both indoor and outdoor. 
Our investigation demonstrates that fine tuning architectures with ArcFace loss in
an end-to-end manner outperforms other two losses by approximately $1\sim4\;\%$ in outdoor 
and $1\sim2 \; \%$ in indoor datasets, given certain thresholds, for the 
visual place recognition tasks.
\end{abstract}

\iffalse
\fi

\section{Introduction}\label{sec:intro}
Image retrieval is a problem of searching for a query image in a 
large image database given the visual content of an image. 
Large-scale Visual Place Recognition (VPR) is commonly formulated 
as a subcategory of image retrieval problem to identify images 
which belong to similar places with same positions in robotics 
and autonomous systems \cite{j:VPR_suevey_DL}. It can generally be 
extended to broader areas, including topological mapping, loop closure 
and drift removal in geometric mapping and learning scene dynamics 
for long-term localization and mapping. Long-term operations in 
environments can cause significant image variations including 
illumination changes, occlusion and scene dynamics \cite{c:VPR_glob2loc}.

The recognition robustness of VPR systems depends on whether or not the matched 
images are taken at the same places in the real world given a certain threshold. 
Therefore, the retrieval performance is highly correlated with feature 
representation and similarity measurement \cite{j:DL_CBIR}. 
One of the crucial challenges is to extract meaningful information of raw images 
in order to mitigate the semantic gap of low level image representation perceived 
by machines and high level semantic concept perceived by human \cite{c:DL_CBIR_STUDY}.

Global feature descriptors are commonly utilized to obtain high image 
retrieval performance with compact image representations. 
Before deep learning era in computer vision, they were mainly developed 
by aggregating hand-crafted local descriptors. Deep Convolutional Neural 
Networks (DCNNs) are now the core of the most state-of-the-art computer 
vision techniques in wide variety of tasks, including image classification, 
object detection and image segmentation \cite{j:recent_adv_cnn}.

Recent advances of the generic descriptors extracted from DCNNs can learn 
discriminatory and human-level representation to provide high-level visual
content of the image patterns \cite{c:conv_nural_code_for_IR}. 
These architectures can potentially learn features at multiple abstraction 
levels to map large raw sensory input data to the output, without relying
on human-crafted features \cite{j:cbir_with_DLprocess}. The DCNN-based 
global features are high performing descriptors trained with ranking-based 
or classification losses \cite{j:delg}.

\begin{figure}[htbp]
%\centerline{\includegraphics[width=1.0\linewidth,height=90pt]{mmsp_teaser.png}}
\centerline{\includegraphics[width=260pt,height=95pt]{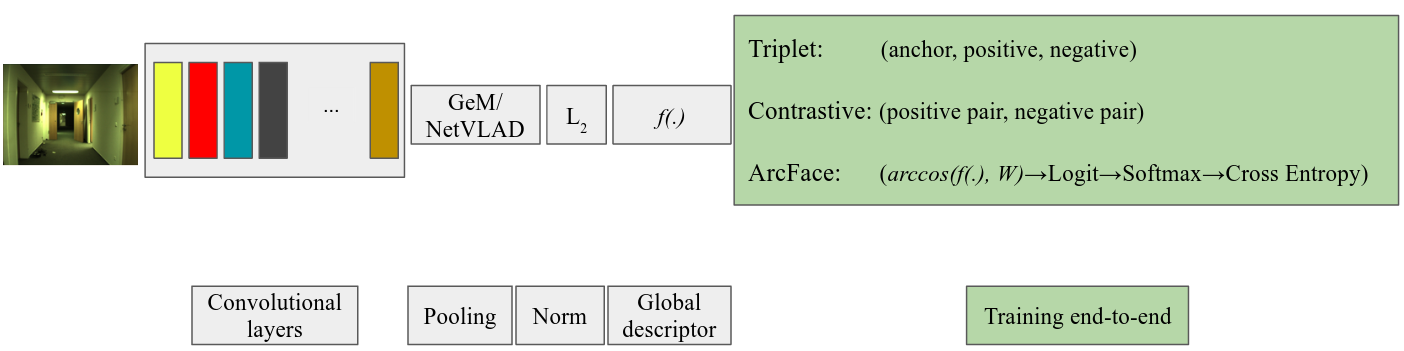}}
\caption{The architecture of our DCNN with the triplet, contrastive and ArcFace losses used for end-to-end training. A single vector global descriptor $f(.)$ is extracted to represent an image.}
\label{fig:teaser}
\end{figure}

\iffalse
%###################################################################
% Original
%###################################################################
Throughout the past few years, several DCNNs have been 
designed to address image classification and objection detection. 
On the one hand, these architectures have robust performance to tackle 
complex computer vision tasks with more large-scale images thanks to the 
larger training datasets and availability of more advanced computational 
resources \cite{j:recent_adv_cnn}. On the other hand, training DCNNs is 
more demanding and requires updating the distribution of layer inputs 
since parameters of previous layers change accordingly. It essentially 
causes slower training process with smaller learning rates and more 
careful parameter initialization.
\fi

%###################################################################
% New paragraph for contribution
%###################################################################
Throughout the past few years, several DCNNs have been 
designed to address image classification and objection detection. 
In this paper, our contribution is to conduct an all-embracing study 
on supervised fine-tuning of such DCNNs with two state-of-the-art pooling 
layers with learnable parameters, one for image retreival, i.e., GeM, and one for 
place recognition, i.e., NetVLAD, in an end-to-end manner for the VPR task on real world
datasets. In our investigation, we report and analyze the performance of 
deep learned global features which are trained with three different loss functions 
in terms of the fraction of correct matches of query images compared to the 
reference images.

The rest of this paper is organized as follows. Section \ref{sec:RW} 
briefly discusses the related work. In section \ref{sec:baseline_methods}, 
we provide the baseline methods along with section \ref{sec:datasets} 
which explains two real-world indoor and outdoor datasets to address 
the VPR task. In section \ref{sec:res}, we show the experimental 
results and we conclude the paper in section \ref{sec:conclusion}.

\section{Related Work}\label{sec:RW}

Before the rise and dominance of DCNNs, VPR methods utilized 
conventional hand-crafted features such as local and global 
image descriptors, to match query images with the reference database 
\cite{c:seqSLAM, j:fabmap2.0, c:VLAD}.
However, deep features have shown more robust 
performance than handcrafted features to address the geometric 
transformations and illumination changes. 

DCNN-based VPR methods can be divided into two main categories.
1) Methods with pre-trained DCNN models, utilized as feature extractor to 
construct image representation to measure the image similarity. 
2) Methods with fine-tune models on specific VPR datasets or new 
architecture to improve the recognition performance.

Chen et al. \cite{j:cnn_based_PR} introduced the first off-the-shelf 
DCNN-based VPR system with features extracted from a pre-trained 
Overfeat \cite{a:overfeat} model in challenging conditions. 
S{\"u}nderhauf et al. \cite{c:on_the_performance} evaluated the 
intermediate \textit{conv3} layer of AlexNet, primarily trained 
for image classification, as holistic image representation to match places 
across condition changes between query and reference database.
Chen et al. \cite{c:DL_feat_at_scale} trained AmosNet and 
HybridNet on a large-scale dataset to address 
the appearance and viewpoint variations. Lopez et al. \cite{j:lopez} 
fine-tuned AlexNet pre-trained architecture on ImageNet dataset for VPR 
with triplets of images containing original image stacked with similar and dissimilar pairs. 
Feature post-processing methods, e.g., feature augmentation and standardization have shown 
improvement compared to using raw holistic DCNN-descriptors 
\cite{c:dont_look_back, c:cnn_img_rep_LC, c:feat_post_pca_whitening}.

DCNN-based VPR methods can learn robust feature mapping to enable image comparison 
using similarity measures such as Euclidean distance. 
Single feedforward pass methods take the whole image as an 
input followed by pooling or aggregation on the raw features to 
design more global discriminative features in which the whole image 
is described by as single feature vector, i.e., global descriptor.

A number of architectures have been proposed to address the image 
representation with global descriptor: MAC~\cite{c:mac_Azizpour}, 
SPoC~\cite{c:spoc}, CroW~\cite{c:CroW}, GeM~\cite{j:Radenovic-TPAMI19}, 
R-MAC~\cite{a:RMAC}, modified~R-MAC \cite{j:RMAC_fine_tune}. Moreover, 
there are DCNNs especially trained for VPR to retreive either global 
image descriptors, e.g., NetVLAD \cite{j:netVLAD} or local features, 
e.g., DELF \cite{c:delf}. 

In this paper, we selected and fine-tuned two state-of-the-art methods 
for global descriptors, e.g., GeM and NetVLAD 
which are trained in an end-to-end manner for both indoor and outdoor VPR datasets.

\section{Baseline Methods}\label{sec:baseline_methods}

\subsection{Pooling Layer}\label{ssec:pooling_ly}
Adding pooling layer after the convolutional layer 
is a common pattern for layer ordering of DCNNs 
to create a new set of the same number of pooled feature maps.
In what follows, we provide a brief summary of GeM and NetVLAD
pooling layers, utilized as pooling layers in this paper:
\paragraph{GeM}
Radenovi\'c et al.~\cite{c:Radenovic-ECCV16, j:Radenovic-TPAMI19}
adopt the siamese architecture for training. It is trained using positive and negative 
image pairs and the loss function enforces large distances between negative pairs, 
e.g., images from two distant places and small distances between positive pairs, e.g., 
images from the same place. Feature vectors are global descriptors of the input images 
and pooled over the spatial dimensions. The feature responses are computed from $K$ convolutional
layers $\mathcal{X}_k$ following with max pooling layers that
select the maximum spatial feature response from each layer of MAC:
\begin{equation}
    \vec{f} = \left[f_1~f_2~\ldots~f_K\right],~f_k = \max_{x \in \mathcal{X}_k} x \enspace.
\end{equation}

GeM pooling layer is proposed to modify MAC \cite{c:mac_Azizpour} and SPoC \cite{c:spoc} 
for better performance. This is a pooling layer which takes $\chi$ as an input and produces 
a vector $f = [f_{1}, f_{2}, f_{i},..., f_{K}]^T$ as an output of the pooling process which results in:
\begin{equation}
    f_{i} = \left(\frac{1}{|\chi_{i}|}\sum_{x\in\chi_{i}} x^{p_{i}}\right)^\frac{1}{p_{i}}
\label{eq:GeM}
\end{equation}

MAC and SPoC are special cases of GeM pooling layer depending 
on how pooling parameter $p_{k}$ is derived in which $p_{i}\rightarrow\infty$ 
and $p_{i} = 1$ correspond to max-pooling and average pooling, respectively. 
The GeM feature vector is a single value per feature map and its dimension 
varies depending on different networks, i.e. $K=[256,\;512,\;2048]$. 

\paragraph{NetVLAD}

Function $f$ is defined as the global feature vector for a given image $I$ as $f(I)$. 
This function is used to extract the feature vectors from
the entire {\em reference} database ${I_{i}}$. Then visual search between $f(q)$, e.g.,
{\em query} image and the reference images $f(I_{i})$ is performed using Euclidean
distance $d(q, I_{i})$ and by selecting the top-N matches. NetVLAD is inspired by the conventional
VLAD~\cite{c:VLAD} which uses handcrafted SIFT descriptors~\cite{j:SIFT} and uses VLAD encoding
to form $f(I)$.

To learn the representation end-to-end, NetVLAD contains two main building blocks. 
1) Cropped CNN at the last convolutional layer, identified as a dense descriptor 
with the output size of $H\times W\times D$, correspond to set of $D$-dimensional 
descriptors extracted at $H\times W$ spatial locations. 2) Trainable generalized 
VLAD layer, e.g., Netvlad which pools extracted descriptors into a fixed image 
representation in which its parameters trained via backpropagation.

The original VLAD image representation $V$ is $D\times K$ matrix in which $D$ is 
the dimension of the input local image descriptor ${\vec{x}_{i}}$ and $K$ is the 
number of clusters. It is reshaped in to a vector after $L_2$-normalization and
$(j,k)$ element of $V$ is calculated as follows:
\begin{equation}
    V(j,k) = \sum_{i=1}^N a_k(\vec{x}_i)(x_i(j)-c_k(j)),
    \label{eq:VLAD}
\end{equation}
where $x_i(j)$ and $c_k(j)$ are the $j$th dimensions of the $i$th descriptor and
$k$th cluster centers, respectively. $a_k(\vec{x}_i)=0,1$ indicates whether or 
not descriptor $\vec{x}_i$ belongs to $k$th visual word. Compared to original VLAD, 
Netvlad layer is differentiable thanks to its soft assignment of descriptors to multiple clusters:

\begin{equation}
V(j,k) = \sum_{i=1}^N \frac{e^{w_{k}^Tx_{i}+b_{k}}}{\sum_{k^\prime}e^{w_{k^\prime}^Tx_{i}+b_{k^\prime}}} (x_{i}(j) - c_{k^\prime}(j))
\label{eq:netvlad}
\end{equation}
where ${w_{k}}$, ${b_{k}}$ and ${c_{k}}$ are gradient descent optimized parameters of $k$ clusters.

In our experiments with 64 clusters and $512$-dimensional VGG16 backbone, 
the NetVLAD feature vector dimension becomes $512 \times 64 = 32,768$. 
Arandjelovic~et~al.~\cite{j:netVLAD} used PCA dimensionality reduction 
method as a post-processing stage. However, in our experiment we utilized full size of feature vector. 
Since NetVLAD layer can be easily plugged into any other CNN architecture 
in an end-to-end manner, we investigate its performance with VGG16 and 
ResNet50 backbones and report the results in section \ref{sec:res}. 

\subsection{Loss Function}\label{ssec:lossFCN}
One of the main challenges of feature learning in large-scale 
DCNN-based VPR methods is to design an appropriate loss function 
to improve the the discriminative power and recognition 
ability~\cite{j:loss_survey_ML_DL}. 
We evaluated the loss functions for the VPR task in two categories. 
1) Measure the difference between samples based on the Euclidean 
space distance, e.g., contrastive loss and triplet loss. 
2) Measure the difference between samples based on 
angular space, e.g., ArcFace loss.

\paragraph{Contrastive}
Given the siamese architecture, the training input consists of 
image pairs $(a,b)$ and labels $y(a,b) = \{0,1\}$ declaring whether 
a pair is non-matching ($y=0$) or matching ($y=1$). 
Contrastive loss~\cite{c:contrastiveLoss} acts on 
matching (positive) and non-matching (negative) pairs as follows:
\begin{equation}
    \mathcal{L_C} = \begin{cases}
        {l}(\vec{f}_a, \vec{f}_b) \hbox{ for matching images}\\
        \max \left(0,M-{l}(\vec{f}_a, \vec{f}_b)\right) \hbox{ otherwise}
    \end{cases}
\end{equation}
where ${l(.)}$ is the pair-wise distance and $M$ is the enforced 
minimum margin between the negative pairs.
$\vec{f}_a$ and $\vec{f}_b$ denote the deep feature vectors of images
$I_a$ and $I_b$ computed using the convolutional head of a backbone
network such as AlexNet, VGGNet or ResNet which leads to feature vector 
lengths $K$ of 256, 512 or 2048, respectively.

\paragraph{Triplet}
The idea of triplet ranking loss is two folds: 
1) to obtain training dataset of tuples $(q,\{{p_i}^q\}, \{ {n_j}^q \})$ 
in which for every query image $q$, there exists set of 
positives $\{{p_i}^q\}$ with at least one image matching 
the query and negatives $\{{n_j}^q\}$, 2) to learn an 
image representation $f_\theta$ so that 
$d_\theta(q,{p_i}^q) < d_\theta(q,{n_j}^q), \forall j$ \cite{c:facenet_triplet_loss}.
Accordingly, we utilized supervised ranking loss 
$\mathcal{L}_T$ adopted by DCNN as a sum of individual 
losses for all $n^q_j$ computed as follows:
\begin{equation}
    \small
    \mathcal{L}_T = \sum _j max(\left(|f_\theta(q) - f_\theta(p^q_i)|^2 - |f_\theta(q) - f_\theta(n^q_j)|^2 + \alpha \right),0),
    \label{eq:triplet_ranking_loss}
\end{equation}
where $\alpha$ is the given margin in meter. 
If the margin between the distance to the negative 
image and to the best matching positive is violated, 
the loss is proportional to the degree of violation. 

\paragraph{ArcFace}
Deng et al. \cite{c:arcface} propose an Additive Angular Margin
Loss (ArcFace) to improve the discriminative power of the global 
feature learning by inducing smaller intra-class appearance variations to 
stabilize the training process.

In this paper, we adopt ArcFace loss function with $L_2$-normalized $D$-dimensional
features $x_i\in \mathbb{R}^{D} $ and classifier weights $W\in \mathbb{R}^{D \times n}$, 
followed by scaled softmax normalization and cross-entropy loss for global feature learning 
tailored for the VPR task, computed as follows:
\begin{equation}
\mathcal{L}_{A} = -\frac{1}{N}\sum_{i=1}^N log \frac{e^{s(cos(\theta_{y_{i}}+m))}}{e^{s(cos(\theta_{y_{i}}+m))}+\sum_{j=1, j \neq y_{i}}^n e^{scos\theta_{j}}},
\label{eq:arcface}
\end{equation}
where $N$ is the batch size, $s$ is the feature scale, $m$ is 
the angular margin and $n=\{0,1\}$ is the class label, e.g., 
binary classification. $\theta_{y_i}$ is the computed angle 
between the feature and the ground truth weight. 
The normalization step on features and weights makes the 
predictions only depend on the angle between 
the feature $x_{i}$ and the ground-truth weight $W_{y_i}$. 
The learned embedding features are thus distributed on a 
hyper-sphere with a radius of $s$.
%%%%%%%%%%%%%%%%%%%%%%%%%%%%%%%%%%%%%%%%%%%%%%%%%%%%%%%%%%%%%%%%%%%%%%%%%
\section{Datasets}\label{sec:datasets}
We evaluated the methods on both outdoor and indoor VPR datasets. 
We selected two query sequences with gradually increasing difficulty:
1) Test 01 conditions moderately changed, e.g, time of day or illumination; and
2) Test 02 conditions clearly different from the reference. In the following
we briefly describe the datasets and selection of training, reference and
the three test sequences.

\subsection{Oxford Radar RobotCar}\label{ssec:rc_radar_dataset}
The Oxford Radar RobotCar dataset~\cite{c:rc_radar} contains the ground-truth 
optimized radar odometry data from a Navtech CTS350-X Millimetre-Wave FMCW radar. 
The data acquisition was conducted in January 2019 over 32 traversals in central 
Oxford with a total route of 280 km of urban driving. It addresses a variety of 
conditions including weather, traffic, and lighting alterations. The combination 
of one Point Grey Bumblebee XB3 trinocular stereo and three Point Grey Grasshopper2 
monocular cameras provide a 360 degree visual coverage of the scene around the vehicle
platform. The Bumblebee XB3 is a 3-sensor multi-baseline IEEE-1394b stereo camera 
designed for improved flexibility and accuracy. It features 1.3 mega-pixel sensors 
with $66\degree$ HFoV and $1280\times960$ image resolution logged at maximum frame 
rate of 16~Hz. The three monocular Grasshopper2 cameras with fisheye lenses mounted 
on the back of the vehicle are synchronized and logged $1024\times1024$ images at 
average frame rate of 11.1~Hz with $180\degree$ HFoV.
\begin{figure}[t]
    \centering
    \subfigure[]{\includegraphics[width=0.12\textwidth]{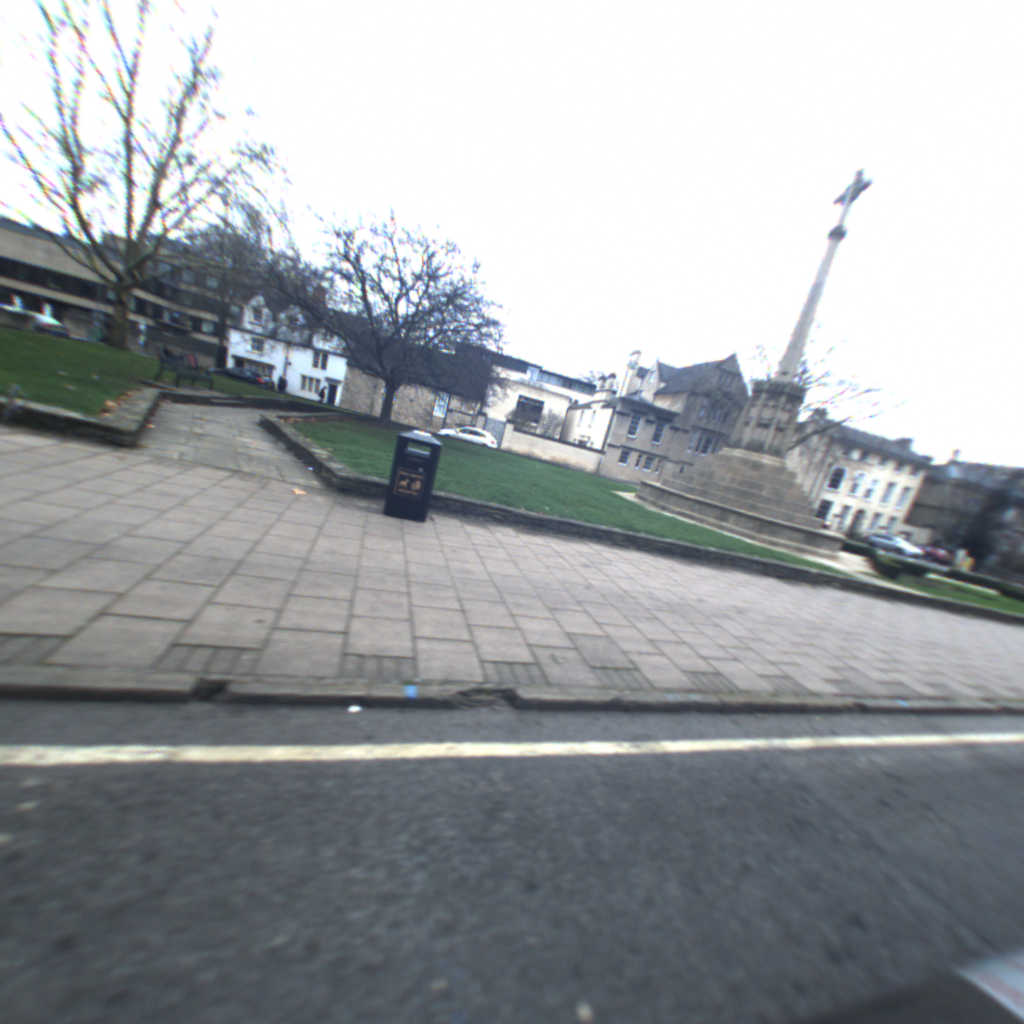}}
    \subfigure[]{\includegraphics[width=0.12\textwidth]{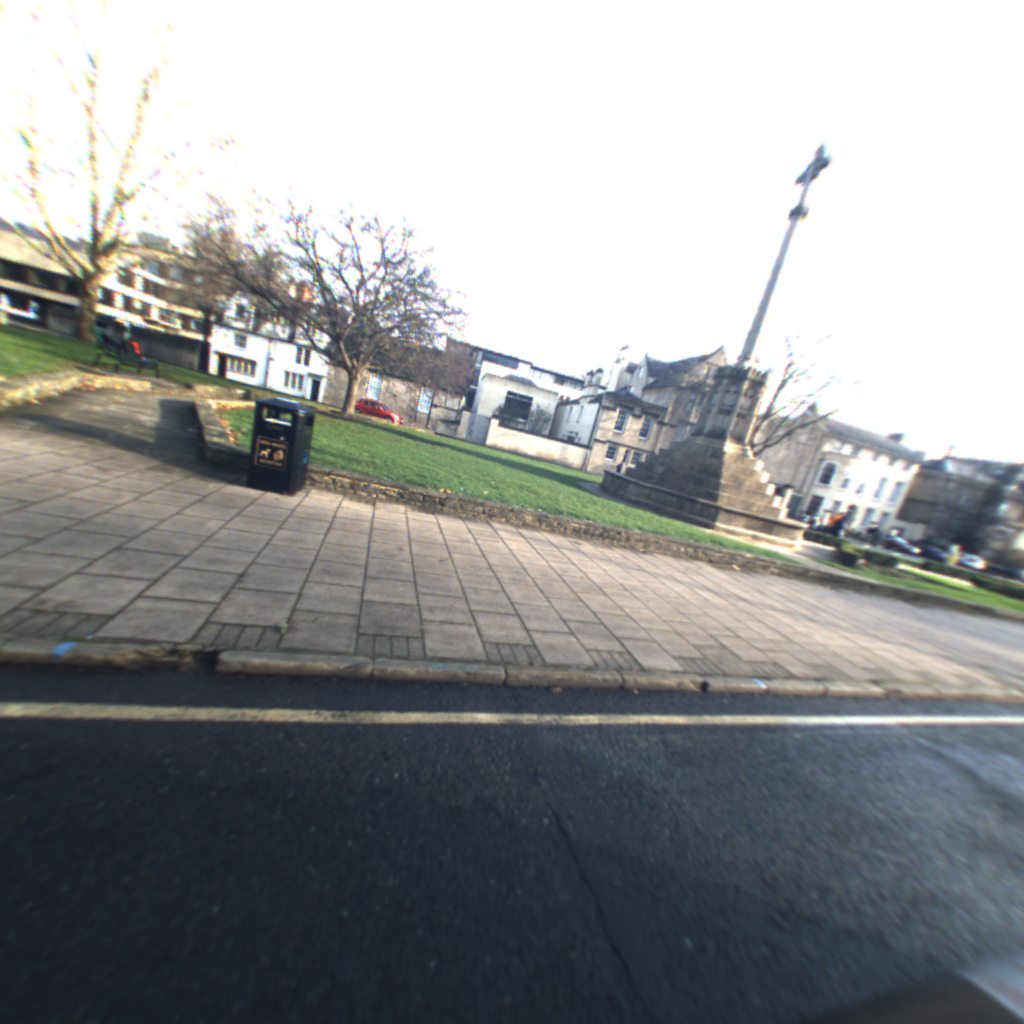}}
    \subfigure[]{\includegraphics[width=0.12\textwidth]{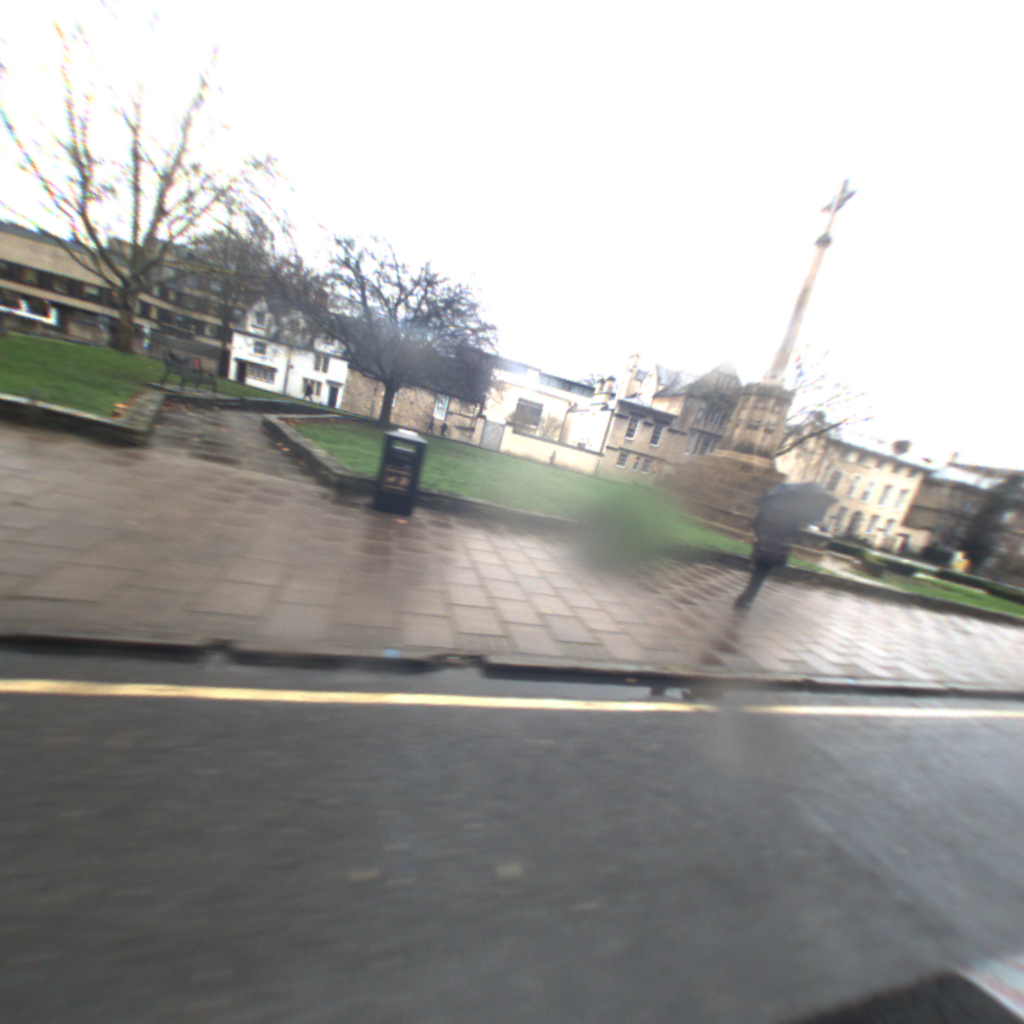}}\\
    \subfigure[]{\includegraphics[width=0.2\textwidth, height=0.185\textwidth]{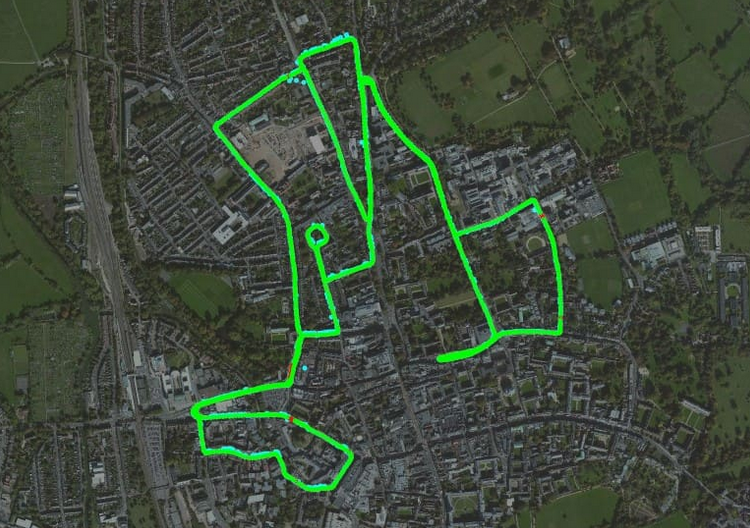}}
    \subfigure[]{\includegraphics[width=0.21\textwidth, height=0.21\textwidth]{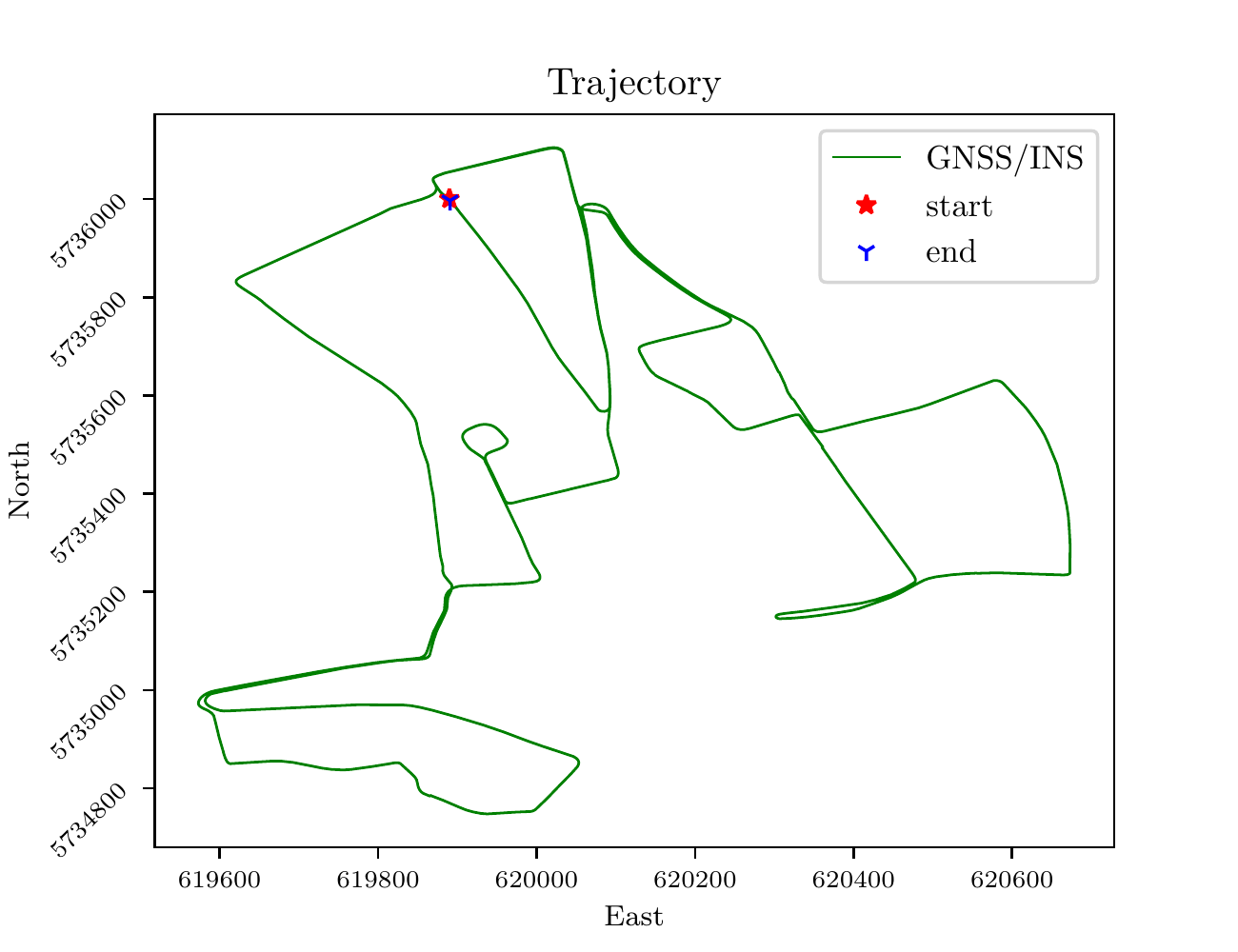}}
    \caption{Examples from the Oxford Radar RobotCar outdoor dataset. Top: Images from the same location in the three selected test sequences: a) Reference: \textit{cloudy} b) Test 01: \textit{sunny} c) Test 02: \textit{rainy} (Grasshopper2 left monocular camera). Bottom: the 19~km route of the test sequences, d) satellite view e) GNSS/INS.}
  \label{fig:robotcar}
\end{figure}
To simplify our experiments we selected images from only one of the cameras, 
the Point Grey Grasshopper2 monocular camera (left), 
despite the fact that using multiple cameras improve the results. 
The selected camera points toward the left side of the
road and thus encodes the stable urban environment 
such as the buildings (Figure~\ref{fig:robotcar}).

From the dataset, we selected sequences for a training set, e.g., 
supervised fine-tuning, a reference against which the
query images from the test sequence are matched and three distinct test sets:
1) the different day but approximately at same time and
2) the different day and different time along with different weather conditions. Table~\ref{tab:robotcar_details} summarizes different sequences used for training, gallery and testing.
\begin{table}[th]
    \caption{The Radar RobotCar outdoor sequences used in our experiments.}
    \label{tab:robotcar_details}
    \centering
    \resizebox{0.9\linewidth}{!}{
    \begin{tabular}{l l l l l}
    \toprule
    {\em Sequence} & {\em Size} & {\em Date} & {\em Start [GMT]} & {\em Condition}\\
    \midrule
    Train       & 37,724 & Jan. 10 2019 & 11:46 & Sunny\\
    Reference   & 36,660 & Jan. 10 2019 & 12:32 & Cloudy\\
    Test 01     & 32,625 & Jan. 11 2019 & 12:26 & Sunny\\
    Test 02     & 28,633 & Jan. 16 2019 & 14:15 & Rainy\\
    \bottomrule
    \end{tabular}}
\end{table}
\subsection{COLD}\label{ssec:cold_dataset}
The CoSy Localization Database (COLD) \cite{j:cold_dataset} presents annotated data sequences acquired using visual and laser range sensors on a mobile platform. The database represents an effort to provide a large-scale, flexible testing environment for evaluating mainly vision-based topological localization and semantic knowledge extraction methods aiming to work on mobile robots in realistic indoor scenarios. The COLD database consists of several video sequences collected in three different indoor laboratory environments located in three different European cities: the Visual Cognitive Systems Laboratory at the University of Ljubljana, Slovenia; the Autonomous Intelligent Systems Laboratory at the University of Freiburg, Germany; and the Language Technology Laboratory at the German Research Center for Artificial Intelligence in Saarbr{\"u}cken, Germany.

Data acquisition were performed using three different robotic platforms (an ActivMedia People Bot, an ActiveMedia Pioneer-3 and an iRobot ATRV-Mini) with two Videre Design MDCS2 digital cameras to obtain perspective and omnidirectional views. Each frame is registered with the associated absolute position recovered using laser and odometry data and annotated with a label representing the corresponding place. The data were collected from a path visiting several rooms and under different illumination conditions, including cloudy, night and sunny.
\begin{figure}[th]
    \centering
    \subfigure[]{\includegraphics[width=0.13\textwidth]{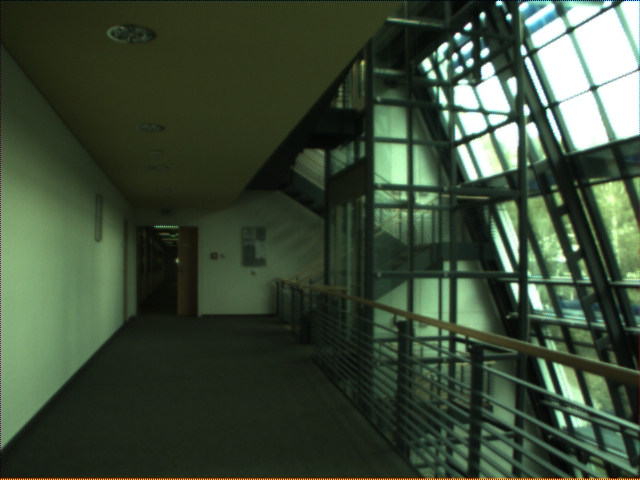}}
    \subfigure[]{\includegraphics[width=0.13\textwidth]{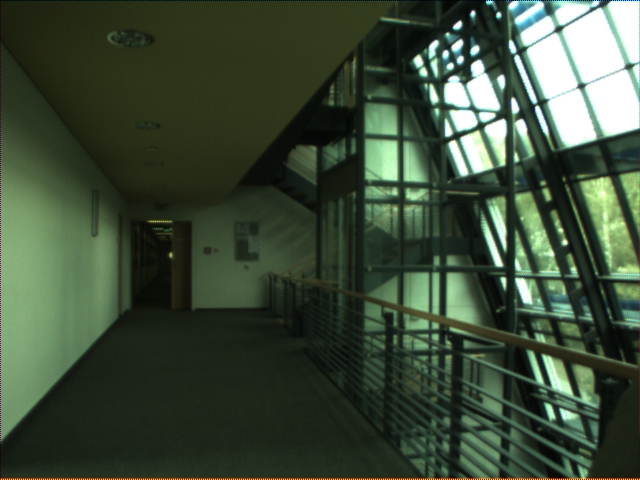}}
    \subfigure[]{\includegraphics[width=0.13\textwidth]{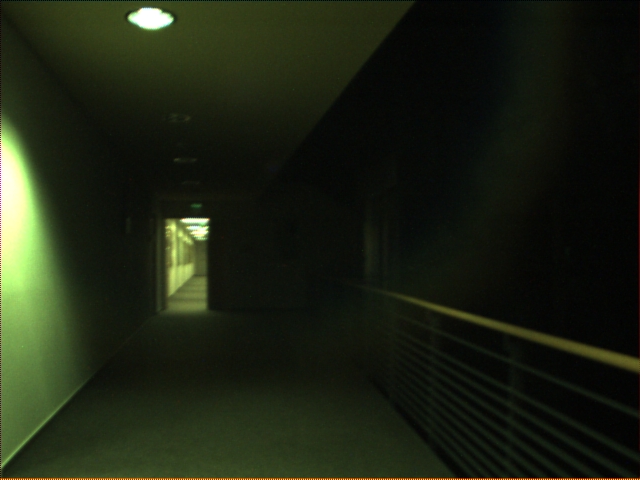}}\\
    \subfigure[]{\includegraphics[width=0.21\textwidth, height=0.35\textwidth]{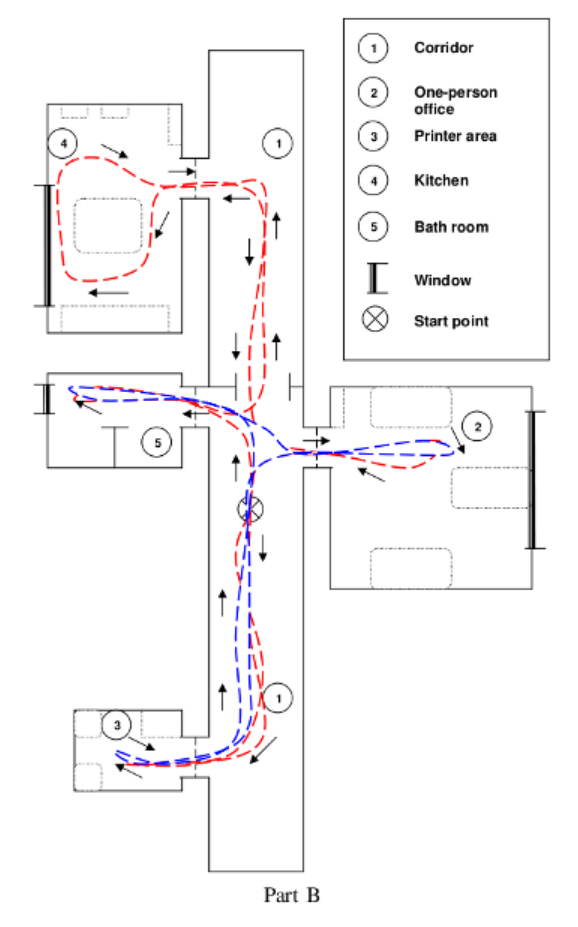}}
    \subfigure[]{\includegraphics[width=0.22\textwidth, height=0.39\textwidth]{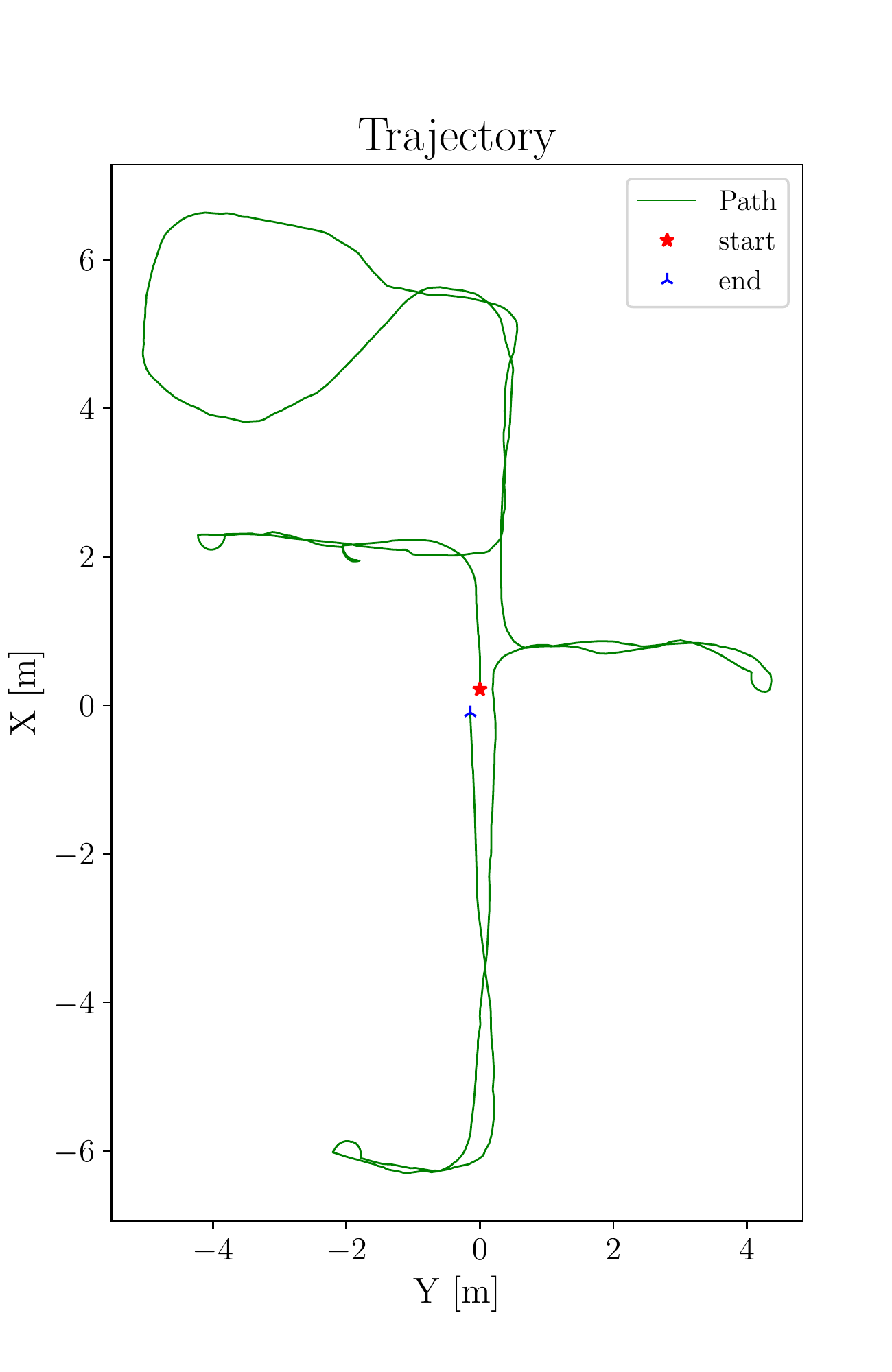}}
    \caption{Examples from the COLD indoor database~\cite{j:cold_dataset}. Images of same location in different sequences: a) Reference: Cloudy-seq1 b) Test 01: Cloudy-seq2, c) Test 02: Night-seq3, d) Map view of the lab: blue dashes: standard path, \textit{i.e.}, short path, red dashes: extended path, arrows indicate the direction of driving the robot and e) Robot path of approximately 50~m.}
    \label{fig:cold_dataset}
\end{figure}
%
%\textcolor{red}{Farid: could you summarize the selected sequences in Table similar to Oxford (Table I)?}
For our experiments, we selected the extended, \textit{e.g.}, 
long path on Map B of Saarbr{\"u}cken laboratory. 
The training sequence is Sunny-seq3, gallery sequence is Cloudy-seq1, 
and the three test sequences are 
1) Sunny-seq1, 
2) Cloudy-seq2 and 
3) Night-seq3. See Figure~\ref{fig:cold_dataset} for examples. 
We used the captured images acquired using the monocular center camera 
form this setup. Table~\ref{tab:cold_details} summarizes different 
sequences used for training, reference and testing.
\begin{table}[ht]
    \caption{The COLD indoor sequences used in our experiments.}
    \label{tab:cold_details}
    \centering
    \resizebox{0.9\linewidth}{!}{
    \begin{tabular}{l r l r l}
    \toprule
    {\em Sequence} & {\em Size} & {\em Date} & {\em Start [GMT]} & {\em Condition} \\
    \midrule
    Train       & 1036 & July 7 2006 & 14:59 & Sunny\\
    Reference   & 1371 & July 7 2006 & 17:05 & Cloudy\\
    Test 01     & 1021 & July 7 2006 & 18:59 & Cloudy\\
    Test 02     & 970  & July 7 2006 & 20:34 & Night\\
    \bottomrule
    \end{tabular}}
\end{table}
\section{Experimental Results}\label{sec:res}
We illustrate the utility of two DCNN architectures with GeM and NetVLAD 
pooling layers for VPR taks, trained and fine-tuned for the indoor and 
outdoor VPR datasets with three different loss functions.
The experiments aim to address the following research questions:
1) what is the accuracy level of place recognition?
2) what is the impact of fine tuning with different 
loss functions, given Euclidean space distance or angular space? 

Given the reference database and query images, we extract 
the sparse feature vectors and compute the similarity 
between each query image and the reference database 
using $L_2$ distance metric. Similar reference feature 
vectors have the lowest $L_2$ distance with the query. 
Similar to \cite{c:dl_off_the_shelf_comparison_own}, we 
calculate Fraction of Correct Matches (FCM) as follows:
\begin{equation}
FCM = \frac{Correct\;Matches}{\sum Query\;Images}\label{eq:fcm} \times 100,
\end{equation}
which corresponds to the proportion of the query images which 
are correctly matched within a certain accuracy threshold $\tau$.

\subsection{Oxford Radar RobotCar}\label{ssec:res_rc}
Depending on which loss function we utilized to fine-tune the 
DCNN-based VPR methods for Oxford Radar RobotCar dataset, 
we report the FCM for two test queries. For this dataset, 
we report the results within meter accuracy threshold, e.g., $\tau = 2\sim25 \; m$.
\begin{table}[tb]
    \caption{FCM for Oxford Radar RobotCar dataset, given trained and fine-tuned model with \textbf{Triplet}, \textbf{Contrastive} and \textbf{ArcFace} loss function.}
    \label{tab:rc_results}
    \centering
    \resizebox{0.99\linewidth}{!}{
    \begin{tabular}{llrrrr}
    \toprule
    {\em Method} & BB & $\tau=25~m$ & $\tau=10~m$ & $\tau=5~m$ & $\tau=2~m$\\
    \midrule
    & & \multicolumn{3}{c}{TRIPLET}\\
    \multicolumn{2}{c}{\textit{Test 01 (diff. day, same time)}}\\
         GeM~\cite{j:Radenovic-TPAMI19}     & VGG16    &      90.89  &      89.36  &      60.11  &      42.21\\ 
                                            & ResNet50 &      95.18  &      93.40  &      61.61  &      47.88\\ 
         NetVLAD~\cite{j:netVLAD}           & VGG16    &      56.59  &      48.89  &      38.42  &      17.01\\
                                            & ResNet50 &      57.82  &      51.60  &      42.13  &      23.83\\ 
    \multicolumn{2}{c}{\textit{Test 02 (diff. day and time)}}\\
         GeM~\cite{j:Radenovic-TPAMI19}     & VGG16    &      89.09  &      86.63  &      82.92  &      62.53\\ 
                                            & ResNet50 &      91.74  &      88.92  &      84.20  &      64.86\\ 
         NetVLAD~\cite{j:netVLAD}           & VGG16    &      38.48  &      31.61  &      24.55  &      24.07\\
                                            & ResNet50 &      47.69  &      42.94  &      37.72  &      22.23\\ 
    \midrule
    & & \multicolumn{3}{c}{CONTRASTIVE}\\
    \multicolumn{2}{c}{\textit{Test 01 (diff. day, same time)}}\\
         GeM~\cite{j:Radenovic-TPAMI19}     & VGG16    &      72.69  &      70.30  &      64.33  &      31.92\\ 
                                            & ResNet50 &      71.84  &      68.41  &      60.71  &      32.71\\ 
         NetVLAD~\cite{j:netVLAD}           & VGG16    &      47.30  &      42.44  &      33.90  &      14.03\\
                                            & ResNet50 &      50.02  &      46.67  &      34.86  &      14.19\\ 
    \multicolumn{2}{c}{\textit{Test 02 (diff. day and time)}}\\
         GeM~\cite{j:Radenovic-TPAMI19}     & VGG16    &      60.38  &      58.94  &      56.05  &      42.78\\ 
                                            & ResNet50 &      56.88  &      52.51  &      48.39  &      33.87\\ 
         NetVLAD~\cite{j:netVLAD}           & VGG16    &      30.94  &      25.53  &      20.69  &      11.13\\
                                            & ResNet50 &      41.94  &      38.33  &      32.09  &      18.30\\ 
    \midrule
    & & \multicolumn{3}{c}{ARCFACE}\\
    \multicolumn{2}{c}{\textit{Test 01 (diff. day, same time)}}\\
         GeM~\cite{j:Radenovic-TPAMI19}     & VGG16    &      91.37  &      89.37  &      66.16  &      42.02\\ 
                                            & ResNet50 &  {\bf 96.51} & {\bf 94.08}  & {\bf 68.91} &  { \bf48.09}\\ 
         NetVLAD~\cite{j:netVLAD}           & VGG16    &      36.11  &      25.53  &      22.40  &       9.65\\
                                            & ResNet50 &      70.54  &      63.35  &      52.42  &      23.83\\ 
    \multicolumn{2}{c}{\textit{Test 02 (diff. day and time)}}\\
         GeM~\cite{j:Radenovic-TPAMI19}     & VGG16    &      89.64  &      86.63  &      82.83  &      62.42\\ 
                                            & ResNet50 & {\bf 92.00} & {\bf 89.00} & {\bf 84.62} & {\bf 65.08}\\ 
         NetVLAD~\cite{j:netVLAD}           & VGG16    &      33.58  &      28.05  &      23.12  &      13.36\\
                                            & ResNet50 &      49.68  &      44.46  &      38.07  &      22.23\\ 
    \bottomrule
    \end{tabular}
    }
\end{table}
Table \ref{tab:rc_results} indicate that ArcFace loss function
outperforms the triplet and contrastive losses by approximately
$1\sim4 \; \%$ when utilized in training time. Furthermore, supervised 
fine-tuning of the DCNN with GeM pooling layer demonstrates higher 
robustness in finding the correct matches for queries in the
reference dataset, e.g., higher FCM.

\subsection{COLD}\label{ssec:res_cold}
In this section, we present FCM results for the indoor COLD database, 
given the trained and fine-tuned models with GeM and NetVLAD pooling 
layers and Tiplet, Contrastive and ArcFace loss functions. For indoor 
dataset, we report the results within centimeter accuracy threshold, 
e.g., $\tau = 25\sim100 \; cm$.
\begin{table}[tb]
    \caption{FCM for COLD database, given trained and fine-tuned model with \textbf{Triplet}, \textbf{Contrastive} and \textbf{ArcFace} loss function.}
    \label{tab:cold_results}
    \centering
    \resizebox{0.99\linewidth}{!}{
    \begin{tabular}{llrrrr}
    \toprule
    {\em Method} & BB & $\tau=100~cm$ & $\tau=75~cm$ & $\tau=50~cm$ & $\tau=25~cm$\\
    \midrule
    & & \multicolumn{3}{c}{TRIPLET}\\
    \multicolumn{2}{c}{\textit{Test 01 (cloudy)}}\\
         GeM~\cite{j:Radenovic-TPAMI19}     & VGG16    &    94.32       &       90.11       &   82.66           & 46.33\\ 
                                            & ResNet50 &    93.18       &       94.03       &   84.62           & 46.52\\
         NetVLAD~\cite{j:netVLAD}           & VGG16    &    90.60       &       85.80       &   75.51           & 43.98\\
                                            & ResNet50 &    92.65       &       89.72       &   79.53           & 45.05\\ 
    %\midrule
    \multicolumn{2}{c}{\textit{Test 02 (night)}}\\
         GeM~\cite{j:Radenovic-TPAMI19}     & VGG16    &    84.54       &        82.47      &   75.26           & 43.71\\ 
                                            & ResNet50 &    90.10       &        85.94      &   80.21           & 46.72\\
         NetVLAD~\cite{j:netVLAD}           & VGG16    &    75.88       &        74.33      &   65.26           & 42.99\\
                                            & ResNet50 &    78.25       &        77.32      &   67.73           & 42.68\\ 
    \midrule
    & & \multicolumn{3}{c}{CONTRASTIVE}\\
    \multicolumn{2}{c}{\textit{Test 01 (cloudy)}}\\
         GeM~\cite{j:Radenovic-TPAMI19}     & VGG16    &    89.23       &        86.58      &   79.33           & 43.19\\ 
                                            & ResNet50 &    94.52       &        91.77      &   {\bf 85.01}     & 47.40\\
         NetVLAD~\cite{j:netVLAD}           & VGG16    &    88.54       &        83.64      &   74.24           & 42.21\\
                                            & ResNet50 &    90.40       &        87.46      &   76.49           & 45.05\\ 
    \multicolumn{2}{c}{\textit{Test 02 (night)}}\\
         GeM~\cite{j:Radenovic-TPAMI19}     & VGG16    &    80.82       &        78.35      &   71.86           & 44.12\\ 
                                            & ResNet50 &    82.06       &        79.48      &   73.30           & 44.54\\
         NetVLAD~\cite{j:netVLAD}           & VGG16    &    74.23       &        72.37      &   65.05           & 41.03\\
                                            & ResNet50 &    75.05       &        74.12      &   66.60           & 42.89\\ 
    \midrule
    & & \multicolumn{3}{c}{ARCFACE}\\
    \multicolumn{2}{c}{\textit{Test 01 (cloudy)}}\\
         GeM~\cite{j:Radenovic-TPAMI19}     & VGG16    &    95.10       &        91.48      &   83.25           & {\bf 48.19}\\ 
                                            & ResNet50 &    {\bf 96.53} &   {\bf 94.32}     &   84.75           & 46.33\\
         NetVLAD~\cite{j:netVLAD}           & VGG16    &    89.52       &        82.57      &   72.18           & 43.78\\
                                            & ResNet50 &    91.19       &        87.56      &   78.65           & 45.05\\ 
    \multicolumn{2}{c}{\textit{Test 02 (night)}}\\
         GeM~\cite{j:Radenovic-TPAMI19}     & VGG16    &    86.60       &        83.61      &   76.19           & 44.64\\ 
                                            & ResNet50 &    {\bf 93.45} &   {\bf 86.49}     &   {\bf 81.14}     & {\bf 47.01}\\
         NetVLAD~\cite{j:netVLAD}           & VGG16    &    76.08       &        74.43      &   66.39           & 42.06\\
                                            & ResNet50 &    72.37       &        69.59      &   62.68           & 37.22\\ 
    \bottomrule
    \end{tabular}
    }
\end{table}
According to Table \ref{tab:cold_results}, trained and fine-tuned models 
with ArcFace loss functions results in more robust performance, $1\sim2 \; \%$, for finding
the correct matches in reference database. Moreover, GeM pooling layer 
outperforms NetVLAD when utilized as a global feature vectors. 

We used VGG16 and ResNet50 backbones in our investigations. Although the size of the COLD indoor 
dataset is relatively smaller than Oxford Radar RobotCar and results of ResNet50 
approximately outperforms VGG16, it is computationally more affordable to
utilize the VGG16 architecture as the backbone due to smaller size.

\section{Conclusion}\label{sec:conclusion}

In this paper, we presented a through investigation of DCNN-based VPR methods, primarily trained and fine-tuned in an end-to-end manner with two pooling layers, e.g., GeM and NetVLAD along with three different loss functions, e.g., triplet, contrastive and ArcFace loss functions.

First, the outperforming validity of supervised fine-tuning the DCNN architechtures, purely trained for classification problems, is comprehensively studied for two real world datasets designed for place recognition in variety of challenging conditions, including seasonal and illumination variations.

Second, the results of correctly matched queries with reference database indicate that supervised fine-tuning the DCNN architectures with ArcFace loss outperforms triplet and contrastive losses for indoor and outdoor datasets within a certain accuracy threshold. Our findings also demonstrate that GeM pooling layer outperforms NetVLAD to extract global feature vectors in both indoor and outdoor datasets.

\bibliographystyle{IEEEtran}
\bibliography{my_ref}

\end{document}